\documentclass[conference]{IEEEtran}
\IEEEoverridecommandlockouts
\usepackage{cite}
\usepackage{amsmath,amssymb,amsfonts}
\usepackage{xparse}
\usepackage{algorithmic}
\usepackage{graphicx}
\usepackage{textcomp}
\usepackage{xcolor}
\usepackage{hyperref}
\usepackage{float}
\usepackage{subfigure}
\def\BibTeX{{\rm B\kern-.05em{\sc i\kern-.025em b}\kern-.08em
    T\kern-.1667em\lower.7ex\hbox{E}\kern-.125emX}}
\begin{document}

\title{POSE : \\ Pose estimation Of virtual Sync Exhibit system}

\author{\IEEEauthorblockN{Hao-Tang Tsui*}
\IEEEauthorblockA{\textit{College of ECE} \\
\textit{
\small{National Yang Ming Chiao Tung University}
}\\
henrytsui000@gmail.com}
\and
\IEEEauthorblockN{Yu-Rou Tuan*}
\IEEEauthorblockA{\textit{College of ECE} \\
\textit{
\small{National Yang Ming Chiao Tung University}
}\\
yztuan1129@gmail.com}
\and
\IEEEauthorblockN{Jia-You Chen}
\IEEEauthorblockA{\textit{College of ECE} \\
\textit{
\small{National Yang Ming Chiao Tung University}
}\\
justin041510@gmail.com}

}

\maketitle

\begin{abstract}
Our project is a portable MetaVerse implementation, and we use 3D pose estimation with AI to make virtual avatars do synchronized actions and interact with the environment. The motivation is that we find it inconvenient to use joysticks and sensors when playing with fitness rings. In order to replace joysticks and reduce costs, we develop a platform that can control virtual avatars through pose estimation to identify the movements of real people, and we also implement an multi-process to achieve modularization and reduce the overall latency. 
\\

Our Code: https://github.com/henrytsui000/POSE
\\
\end{abstract}


\section{Introduction}

\huge{A}\normalsize{s}
the Wii swept the world and opened the era of home game consoles, the technology of detecting player movements has become increasingly essential. As a pioneer, Wii used infrared light and a three-axis accelerator to detect player movements.\cite{b1}\cite{b2}
Microsoft also developed the kinect, a joystick-free somatosensory device for the Xbox 360. Kinect uses a camera to capture the player's movements,  then processed by Image recognition for captures human joints.\cite{b3}
\par However, its disadvantage is that it requires a large space to capture actions, and the types of games are relatively simple, mostly fitness sports. At present, there are few cheap games on the market that have both real-life motion detection and interaction with the environment. For example, Ring Fit Adventure can make the virtual avatar move in the environment, but the actions that the avatar can do are fixed.
\par Therefore, we want to capture the movements of real people through the webcams and use pose estimation to capture the movement and rotation of joints. Using Panda3D as a platform, we build a virtual reality, allowing the virtual avatar to detect movements and make the virtual avatar make corresponding actions.\cite{b4}
\par
\begin{figure}[h]
\includegraphics[width=\linewidth]{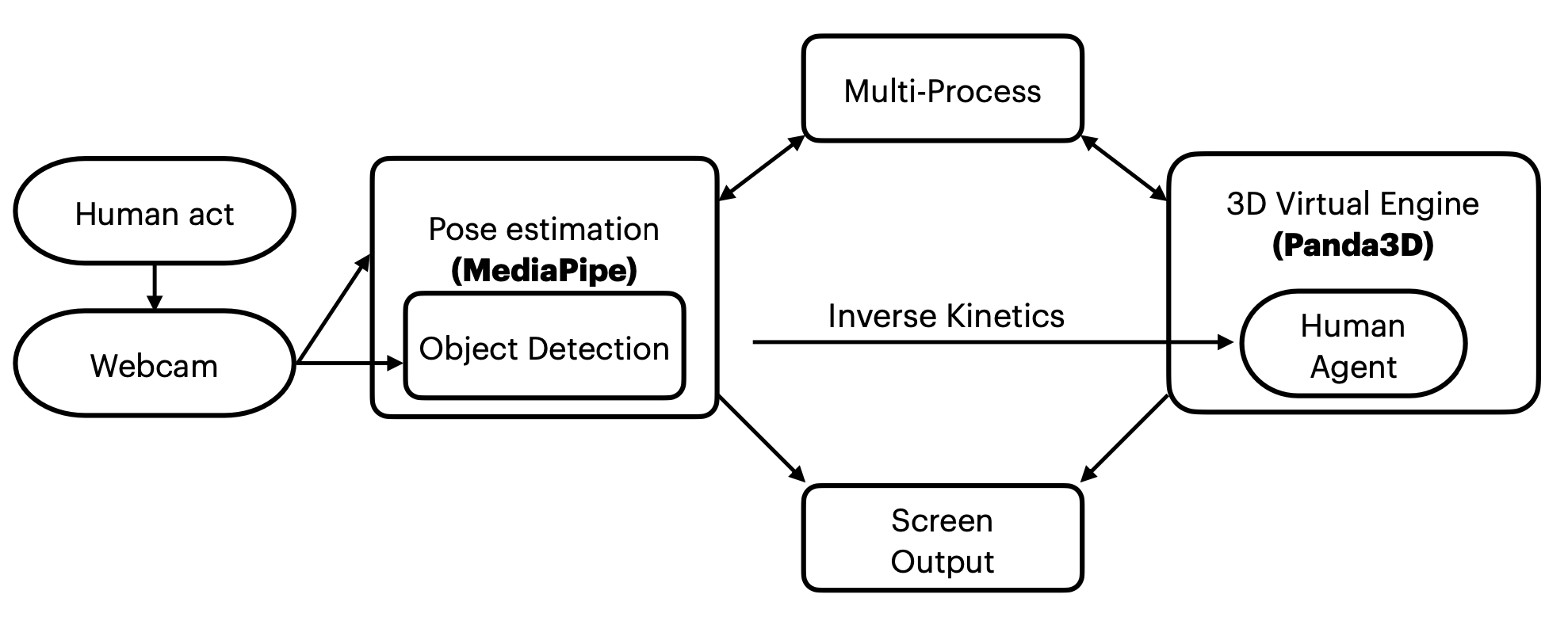}
\caption{Project architecture diagram}
\label{fig:Panel}
\end{figure}

 What's more, in addition to allowing the virtual avatar to move and perform various actions in the virtual reality, we can also interact with the characters in it. Fig.~\ref{fig:Panel} is our project's architecture diagram.

\section{Related Work}

\subsection{Human Pose Estimation}

\begin{figure}[h]
\centering
\subfigure[Keypoints of human pose]{
\label{pose_est1}
\includegraphics[width=0.25\textwidth]{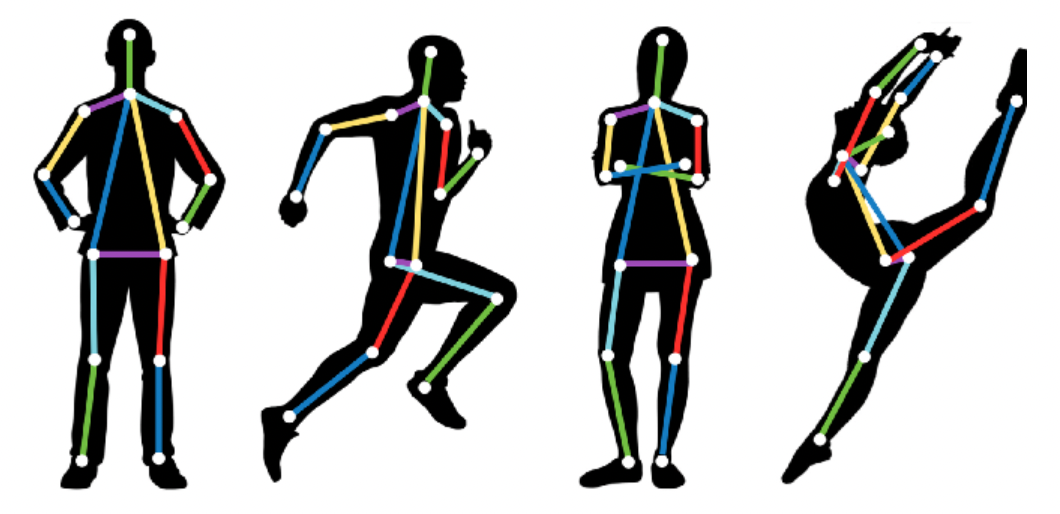}}
\subfigure[Human skeleton after pose estimationP]{
\label{pose_est2}
\includegraphics[width=0.20\textwidth]{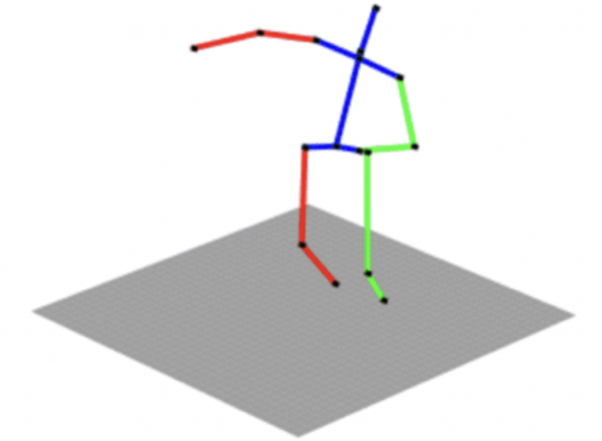}}
\caption{Human Pose Estimation}
\label{Fig.pose}
\end{figure}

Pose estimation is a popular task of using Deep Learning models to estimate poses of human from an image or a video by estimating the spatial locations of body joints. As Fig.~\ref{Fig.pose} shows, Fig.~\ref{pose_est1} is the keypoints diagram of different human poses, and Fig.~\ref{pose_est2} is what the model get after pose estimation. The mechanism of the model is input an image of a person into the model, and the common way is to calculate the heat map of this image, and then detect the positions of multiple joint points of the person. \cite{b5}\cite{b6}


\subsection{Inverse Kinematics}
Inverse kinematics method (IK) is responsible for calculating the joint movement required for the end position. It will calculate the angle of each joint from a terminal joint and the length of each bone to form a robotic arm that touches this distal point.
\par IK operation in human model is shown as Fig.~\ref{avatar}. At present, IK can be realized through quite a variety of algorithms, such as Jacobian method\cite{1401.1488} and machine learning methods that have been gradually proposed in recent years.\cite{2205.10837}

\subsection{3D Virtual Engine}
3D engine is a platform for visualizing virtual avatars and virtual environments.\cite{b9} The common production platforms are mono, unity, panda3d, etc. for developing games. However, different game platforms may also need to be implemented in different programming languages.

\section{Methods}

In this project, we use pose estimation to recognize human actions and send instructions to an agent in the 3D environment. To achieve low latency, real time experience, we design a controller that perform multi-processing computing.
\subsection{3D Pose Estimation}

\subsubsection{Transpose}In the field of Pose estimation, the state-of-the-art model is TransPose, so our pose estimation experiments so far are all carried out using TransPose.\cite{b11} But it is difficult to use 2D image recognition to locate the 3D position, because we will need to calculate the depth position of each joint point, so calculate the 3D position of the joint point position is indispensable. Our solution is to use two lenses to calculate coordinate position, depth, height, joint rotation direction and travel distance of the person's joints, and the distance and relative position between the two lenses are automatically corrected by AprilTags.\cite{b12} 

\subsubsection{MediaPipe}We also consider other lightweight models, such as MediaPipe also provides a 3D Pose Estimaiton model.\cite{1906.08172} The model architecture of MediaPipe is TFLite model, which includes hardware acceleration, so the latency is low. There are a total of three models of different magnitudes, namely lite, full, and heavy. You can choose to use different models according to different Performance requirements and computer computing power.

\subsection{3D Virtual Engine}
We use panda3D, an open source framework for python to implement the 3D environment. Since the cv2 and panda3D integration is complete, it allows us to efficiently create scenes and objects. Thus, it is practical to use panda3D for the work to produce the realistic characters that have joints and can move like real people.
\par Moreover, we can also create human agents in this virtual reality that synchronize the actions of real people, which enable us to interact with the AI or the environment. Fig.~\ref{avatar} shows the first virtual agent model in 3D engine.
\par In addition, we also considered using Unity as the 3D engine at the beginning. The advantage is that its calculation load is relatively small and it has more functions. However, its C\# framework makes it harder for us to use Unity in python. Also, our project is not very large, so faster calculations and more functions are not needed.

\subsection{Inverse Kinematics method}
Because the version of panda3D is the first version, some bugs that cannot be solved by the system often appear when directly controlling the model joints, the coordinates are more complicated, and the Dependencies between joints make the control more inconvenient. So we used the Inverse Kinematics method (IK) to control the movement of the avatar. With IK, the angle of the joint can be found by the end effector. Another advantage of IK is that it can be packaged into functions, and no need to rewrite in other joints.
\par
\begin{figure}[t]
\centering
\subfigure[Ball joint]{
\label{ball}
\includegraphics[width=0.20\textwidth]{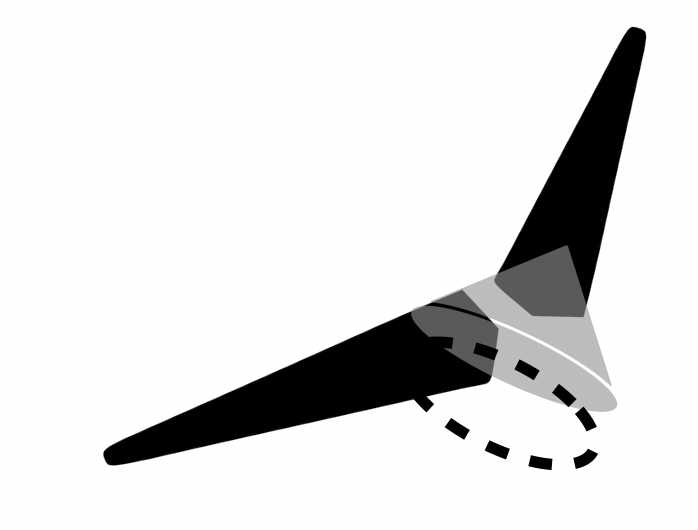}}
\subfigure[Hinge Joint]{
\label{hinge}
\includegraphics[width=0.20\textwidth]{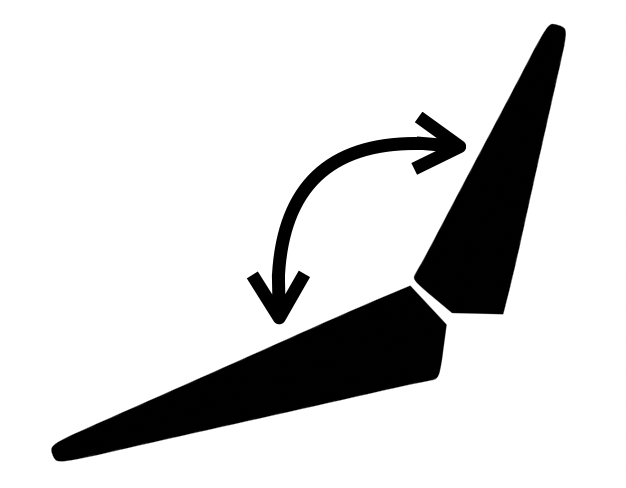}}
\subfigure[Avatar operation of IK]{
\label{avatar}
\includegraphics[width=0.25\textwidth]{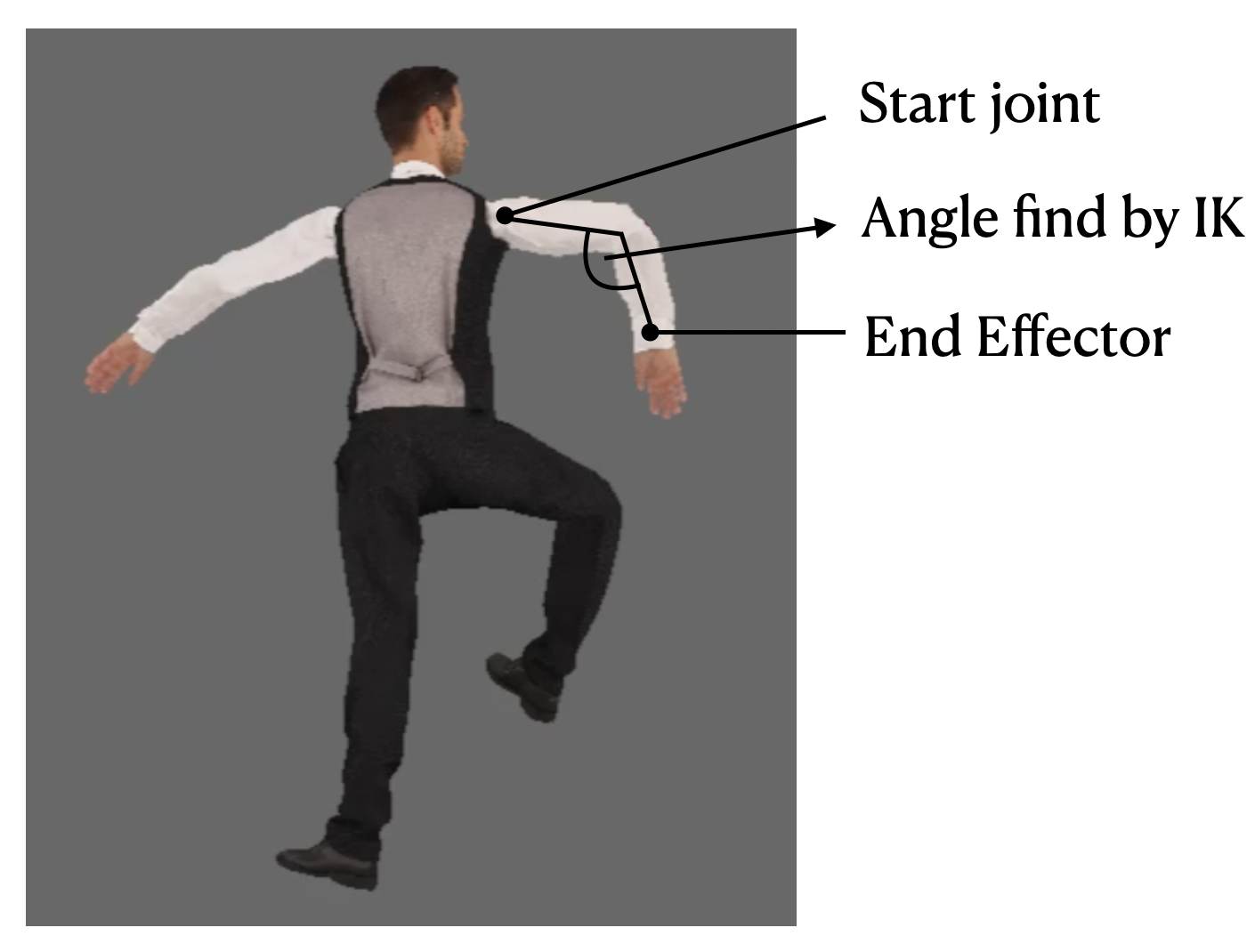}}
\caption{The constraints and the avatar operation of Inverse Kinematics}
\label{IK}
\end{figure}

 In addition, set constraints on the bones of the model in panda3d are also very important, setting constraints can solve the multi-solution of Inverse kinematics and making the movement of the model more realistic. We use two kinds of constraints in total, namely ball constraint and hinge constraint, the mechanisms of both are shown in Fig.~\ref{ball} and Fig.~\ref{hinge}. The limiting angle of the Ball constraint is a cone, and the joint can rotate along the cone. The limit angle of the hinge constraint is a one-dimensional direction. The joint can only rotate along this one-dimensional direction so that the bone can only move along this one-dimensional direction at a time.

\subsection{Math of Convert Coordinate}
\par In fact, mapping joints to the virtual game engine is not as simple as we imagined. There are several reasons:
\begin{itemize}
\item The coordinate system of panda3d is different from the coordinate system obtained by pose estimation $(x, y, z) \rightarrow (x, z, y)$.
\item The joint position in the virtual game engine is usually based on the previous joint, which needs to be rotated, scaled, etc. for the body.
\item We need to find the relative coordinate system to be able to map.
\end{itemize}
\par So we thought of a solution, for each joint, first find the basis vector. And this vector should meet the characteristics of easy identification and long enough length. Name this vector $Base\enspace Vec(\vec{B})$, and the vector we want to rotate as $Joint\enspace Vec(\vec{J})$. For example, when we use the upper left arm as  $\vec{J}$, we will use the right shoulder to shoot to the left shoulder as $\vec{B}$. Even with these reference vectors, we still can't rotate the shoulder enough. At this time, we thought that there is no need to calculate the height information (Y-axis) here. When we finish calculating the x and z axes, we can use the same scaling method to process the y-axis. What we have to do is to find a matrix that can convert $\vec{B}$ to the x-axis unit vector $U_x = (1, 0, 0)$. That is to solve Eq. \ref{eq_pose_rotate}.
\begin{equation}\label{eq_pose_rotate}
\vec{B} \cdot 
\begin{bmatrix}
\cos{\theta_m} & \sin{\theta_m}\\
-\sin{\theta_m} & \cos{\theta_m}
\end{bmatrix}
= \begin{bmatrix}
C \\
0
\end{bmatrix}
\end{equation}

\par What we need to find out is the $\theta_m$ and constant $C$ in Eq. \ref{eq_pose_rotate}. After carefully observing the formula, will find that it is not difficult, because the matrix is a linear rotation matrix, which will not change $||\vec{J}||_2$. So the constant C is the length of $\vec{J}$, but because we plan to finish the calculation on the $x,\enspace z$ plane (overlooking the user) first, and then map to the y-axis, the length is $\sqrt{B_x^2+B_z^2}$. And the calculation of $\theta_m$ is not difficult. In fact, it is the angle between $(B_x, B_z)$ and the x-axis. It can be calculated simply through $\theta_m = \arctan{(\frac{B_z}{B_x})}$ radian.
\par
\begin{figure}[ht]
\centering
\subfigure[Original Joint Vector]{
\label{Joint_original}
\includegraphics[width=0.2\textwidth]{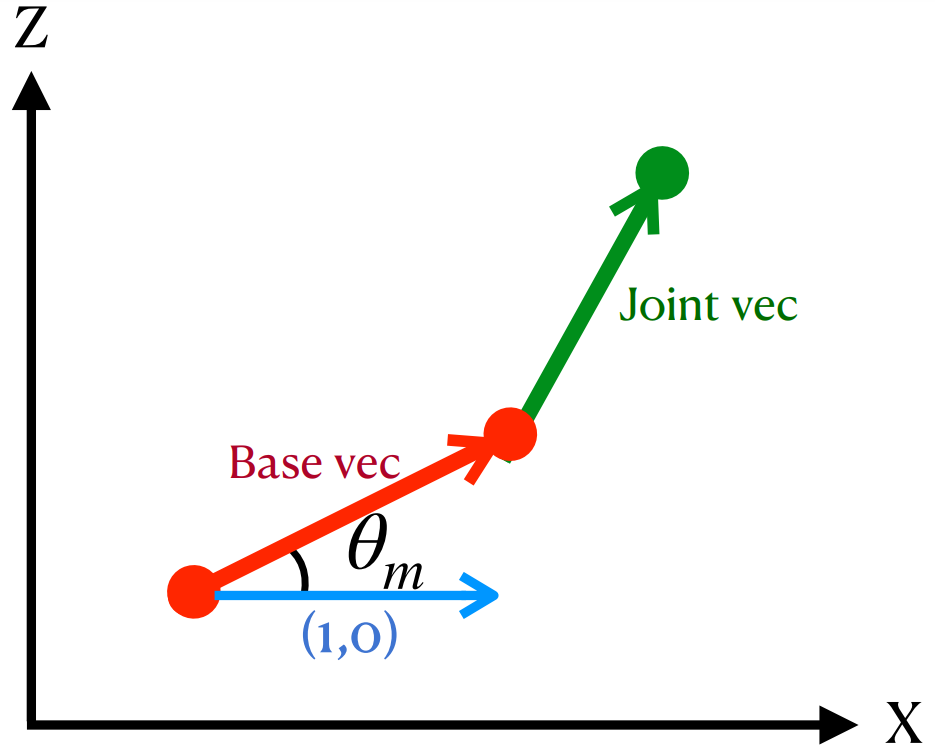}}
\subfigure[Converted Joint Vector]{
\label{Joint_converted}
\includegraphics[width=0.2\textwidth]{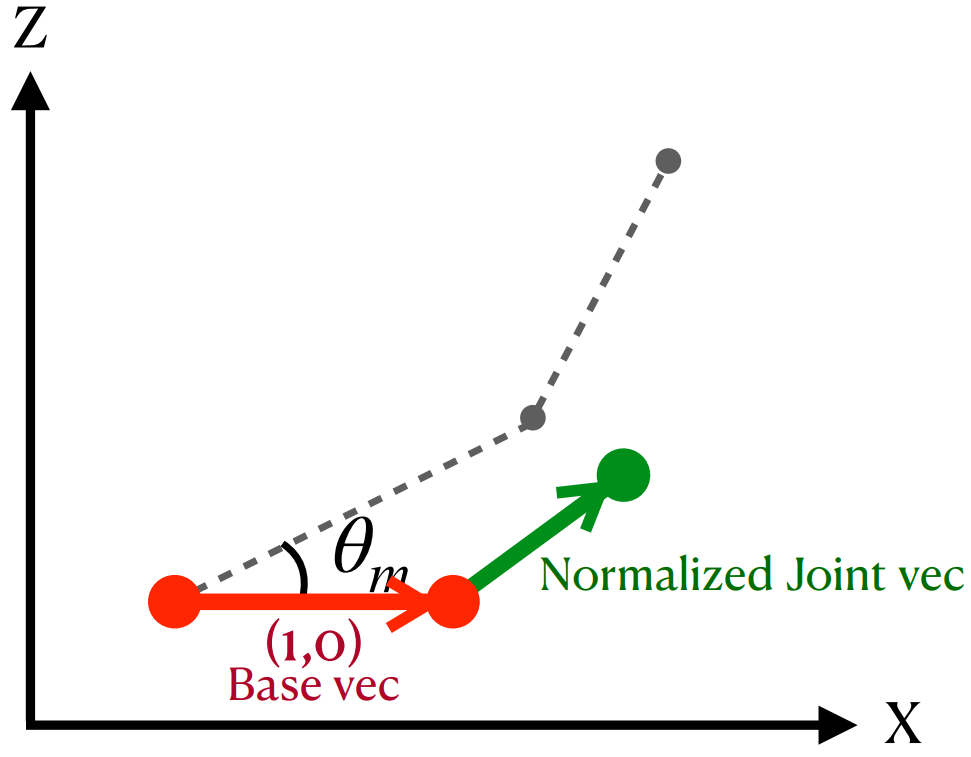}}
\subfigure[Panda3D Joint Vector]{
\label{Joint_avatar}
\includegraphics[width=0.2\textwidth]{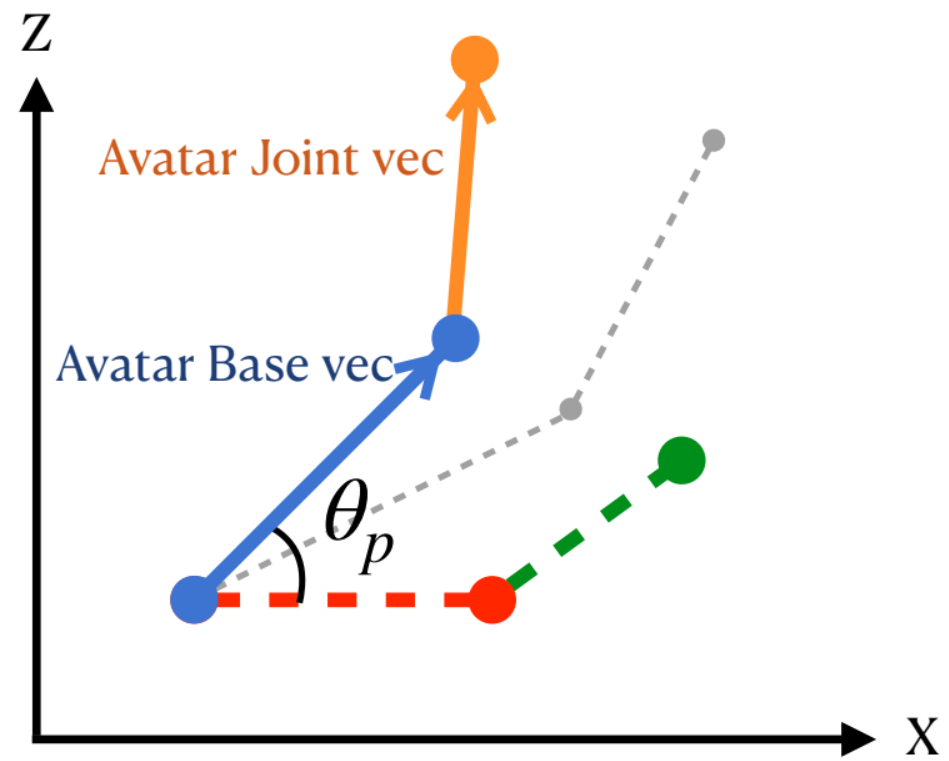}}
\caption{The converting process of joint vectors}
\label{Joint_trans}
\end{figure}

 As a result, we can convert the bird's-eye view in the original picture (Fig. \ref{Joint_original}) to a regular bird's-eye view as in the Fig. \ref{Joint_converted}, that is, the basis vector $\vec{B}$ is the unit vector $U_x$ of the x-axis.
\par Then it is our turn to actually do the vector $\vec{J}$, we use the $\theta_m$ and the constant $C$ we just found, to rotate and scale $\vec{J}$ according to Eq. \ref{eq_pose_rotate_joint}, You can get the normalized joint vector $\vec{\hat{J}}$ under the bird's eye view
\begin{equation}\label{eq_pose_rotate_joint}
\frac{\vec{J}}{C} \cdot
\begin{bmatrix}
\cos{\theta_m} & \sin{\theta_m}\\
-\sin{\theta_m} & \cos{\theta_m}
\end{bmatrix}
= \vec{\hat{J}}
\end{equation}

\par Next is the part of the virtual game engine. After sending the aforementioned $\vec{\hat{J}}$ to the virtual game engine, what we need to do is to restore the normalized vector to the actual joint vector $\vec{J^{\prime}}$. What we have to do is to find the basic vector in the virtual game engine. If we continue the example mentioned above, we will find the vector from the right shoulder to the left shoulder in the game doll ($\vec{B^{\prime}} $). Then we can find a good property Eq. \ref{eq_recover}, and $\theta_p$ is the basis angle between in the virtual game engine $\vec{B^{\prime}}$ and the positive x-axis, which is $\arctan{(\frac{B^{\prime}_z}{B^{\prime}_x})}$, and $C^{\prime}$ is the length of $\vec{B^{\prime}}$ ($\sqrt{B_x^{\prime2}+B_z^{\prime2}}$).

\begin{equation}\label{eq_recover}
\begin{bmatrix}
1 \\
0
\end{bmatrix} \cdot 
\begin{bmatrix}
\cos{\theta_p} & \sin{\theta_p}\\
-\sin{\theta_p} & \cos{\theta_p}
\end{bmatrix}
=  \vec{B^{\prime}}
\end{equation}

Then we reproduce the target vector $\vec{J^{\prime}}$ to the target vector $\vec{\hat{J}}$ according to the previous rotation and scaling methods, that is, apply the target to the puppet in the virtual environment according to the method of \ref{eq_recover_joint} to find the correct joint target point.

\begin{equation}\label{eq_recover_joint}
\vec{\hat{J}} \cdot C^{\prime} 
\begin{bmatrix}
\cos{\theta_p} & \sin{\theta_p}\\
-\sin{\theta_p} & \cos{\theta_p}
\end{bmatrix}
=  \vec{J^{\prime}}
\end{equation}

\subsection{Multi-processing}
\paragraph{multi-processing}
We will design a module controller to schedule each module to maximize time efficiency. The advantage of modularization is that as long as the input and output are defined, it will be much easier to modify the module. You only need to change the content in the module, without changing the overall code.\par
Therefore, modularization of each block, easier division of labor and cooperation. In addition, Synchronize each modulation and reduce latency. Because we can make the result of pose estimation and the operation of Unity engine synchronize through the concept of pipeline, there is no need to wait for each other and increase the delay time. Because if you want to wait for the pose estimation to give the result to the 3D engine, the preprocessing time will be wasted, and if the function of the pose estimation breaks down, the 3D engine will also break down. However, if the module controller is used to arrange the appropriate pipeline, the above problems can be avoided. If the pose estimation breaks down at a certain moment, the 3D engine can use the pose estimation result of the previous second to increase the robustness of the environment.
\paragraph{why not ROS?}
ROS was also under our consideration, but we found that for this topic, light-weight message transmission is required. For example, the video stream only needs to be extracted once, so the videocapture of cv2 can be competent for this matter. Most of the time the transmitted data is transmitted to the enviornment for pose estimation, so I believe that the simple socket server I built can be competent for this, and does not require a high-level ros system.\cite{b10}

\section{Experiments}

\subsection{Program Framework}
\par All of our projects conform to the PEP 8 specification, a set of guidelines for writing clean, readable, and maintainable code in Python. This includes adhering to a consistent indentation style, using meaningful variable names, and following a specific format for docstrings and comments. By following PEP 8, we ensure our code is easy for other developers to understand and adheres to industry best practices.
\par In addition to complying with PEP 8, we also use Git for version control in a clean and organized manner. This includes regularly committing and pushing code changes, using clear and descriptive commit messages, and branching and merging code in a logical and efficient manner. By using Git, we are able to keep track of all changes made to our codebase, easily collaborate with other developers, and revert to previous versions of our code if necessary. We strictly abide by the way Git is used, and do not develop on the master branch. During the function development, we will create meaningful branches and merge them into the main branch after the function development is completed.
\par Overall, our adherence to PEP 8 and use of Git has allowed us to write high-quality, maintainable code that is easy to understand and use, making it a more efficient workflow and more maintainable code.

\subsection{Collaborative Architecture}
\par As mentioned earlier we use Git to maintain development and release versions. We use open source to Github, and its benefits, for example, it has lightweight branches that can be merged and tracked, and can be versioned without server support. It also allows the project to have a complete readme to show usage, improve cooperation efficiency, and reduce misunderstandings in version communication.
\par In addition, we use Gitmoji, which is an emoji memo used when submitting information on Github. It can record the content attributes modified by each commit, which helps to maintain the integrity of the commit without losing its beauty. It is not easy to get lost on the switch.
In this section, we will briefly describe the overall hardware setup, describe the expected difficulties and what we have tried and designed experiments on the accuracy of different models.

\subsection{Environments}

\begin{itemize}
\item GPU:                 RTX 3090
\item Desktop:        Ubuntu 22.04
\item Notebook:     Mac M1 / Win10
\end{itemize}
\par To generalized our project, we specially installed Ubuntu 22.04 and developed our project system on it, because the Ubuntu environment is cleaner and easier to develop than Windows. In order to ensure our systems can run on various types of operating systems. We then deploy our project system on other operating systems, including the popular Mac and Windows operating systems. Also, the GPU we used is RTX 3090. Because the performance of RTX 3090 is currently one of the best in the market.
\par We are able to use the computing power of this GPU to successfully train all models and support the entire project system.
\par We run all our experiments on an Ubuntu environment, including fine-tuning each model, building a panda3D environment, and trying other work, includes building an AI agent to interact with the user. In addition, we also have laptops with Mac M1 system and Windows system to test whether the project can run smoothly on different systems. We expect our final project to be able to run on various devices, whether it is Ubuntu, M1 or Windows operating system. As long as they contain enough computing resources.


\subsection{Ablation Study}
\par
\begin{figure}[H]
\centering
\subfigure[Result of mediapipe]{
\label{mediapipe}
\includegraphics[width=0.20\textwidth]{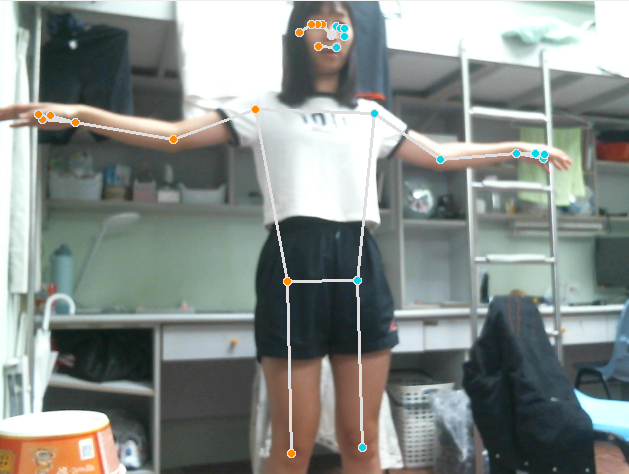}}
\subfigure[Result of Transpose]{
\label{transpose}
\includegraphics[width=0.20\textwidth]{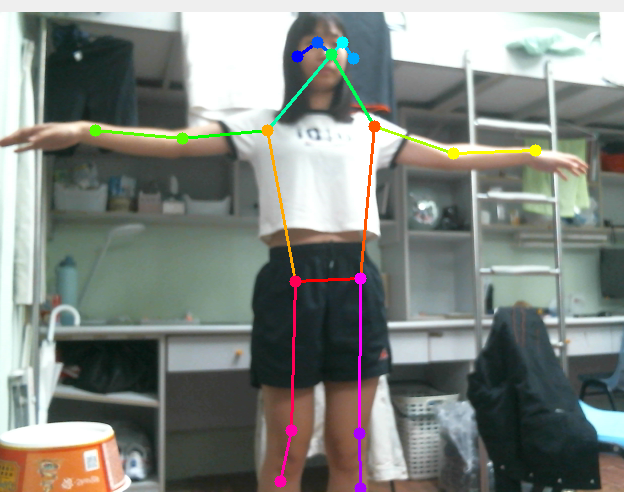}}
\caption{Results of pose estimation model}
\label{Fig.main}
\end{figure}

 Because there are two kind of models, mediapipe and Transpose, we try to make some study about them. The result of the mediapipe (Fig.~\ref{mediapipe}) and Transpose (Fig.~\ref{transpose}) are considered. Although they have similar results, the results of Transpose cannot provide information of the depth without two cameras, while mediapipe can do this. Therefore, we decided to use mediapipe as our final model.
\par In addition, because mediapipe is based on the TFlite architecture, there will be initial hardware acceleration. The BlazePose they provide is composed of some simple convolutions, and there will be no difficulty in deploying it to other operating systems. And there are many models of different sizes, so that we can make the correct trade-off between accuracy and delay according to the computing power of different operating systems in practice. And because there is only simple convolution, the FPS is very high, and the accuracy is very satisfactory. Therefore, we also consider letting him deploy to the FPGA to do calculations.

\section{Conclusion}
\par To sum up, we plan to use the technology of pose estimation to achieve a virtual world effect like a metaverse, so that people without neighbors can talk or even chat face-to-face in the virtual world. At the same time, it can also become a game platform that replaces the joystick, allowing people to lower the threshold for starting games to just a simple computer and lens (that is to say, a laptop can do the job). At the same time, a novel idea is proposed to solve the parallel design between various modules to reduce the model delay.

\subsection{Dual Camera}
\par Although a system requires a camera that is very portable and simple to deploy, it is undeniable that its accuracy is not satisfactory enough. Furthermore, most laptops or webcams are horizontal, that is, there is no way to capture the movement of the feet. However, the recent mac has given us a lot of inspiration. Maybe we can try to use the mobile phone as an extended lens and set it up behind the user. This can make the depth calculation more robust, and also allows the position of the feet to be captured. It's a pity that the special time makes it too late for us to develop, and macOS currently can't use cv2 to extend the mobile phone lens, it can only be used for facetime.

\subsection{Server System}
\par In addition to dual cameras, we are also trying to build servers to allow users to interact in the virtual game engine and realize the actual idea of the Metaverse. Or keep up with the popular concept of decentralization, let the user's computer become one of the operators, and let other users appear in their virtual engine. These are the goals we hope to achieve in the next stage, and we will also open source on GitHub to attract other interested engineers to send pull requests.

\section*{Acknowledgment}

\par Here we are very grateful to the contributors in various fields, from the developers of Wii and switch for me to find business opportunities between them; as well as all researchers who are constantly promoting the field of pose estimation, so that we can directly use their accurate models. As well as panda3d or Unity design team, through this open programming environment, we can quickly create a simple environment. Of course, the teaching assistants and professors of this course gave us valuable opinions and ideas on the topic.

\appendices
\section{Some Pictures for Demonstration}
\par The person in following pictures posing in different poses is Henry Tsui, the author of this project.

\begin{figure}[h]
\centering
\includegraphics[width=0.8\linewidth]{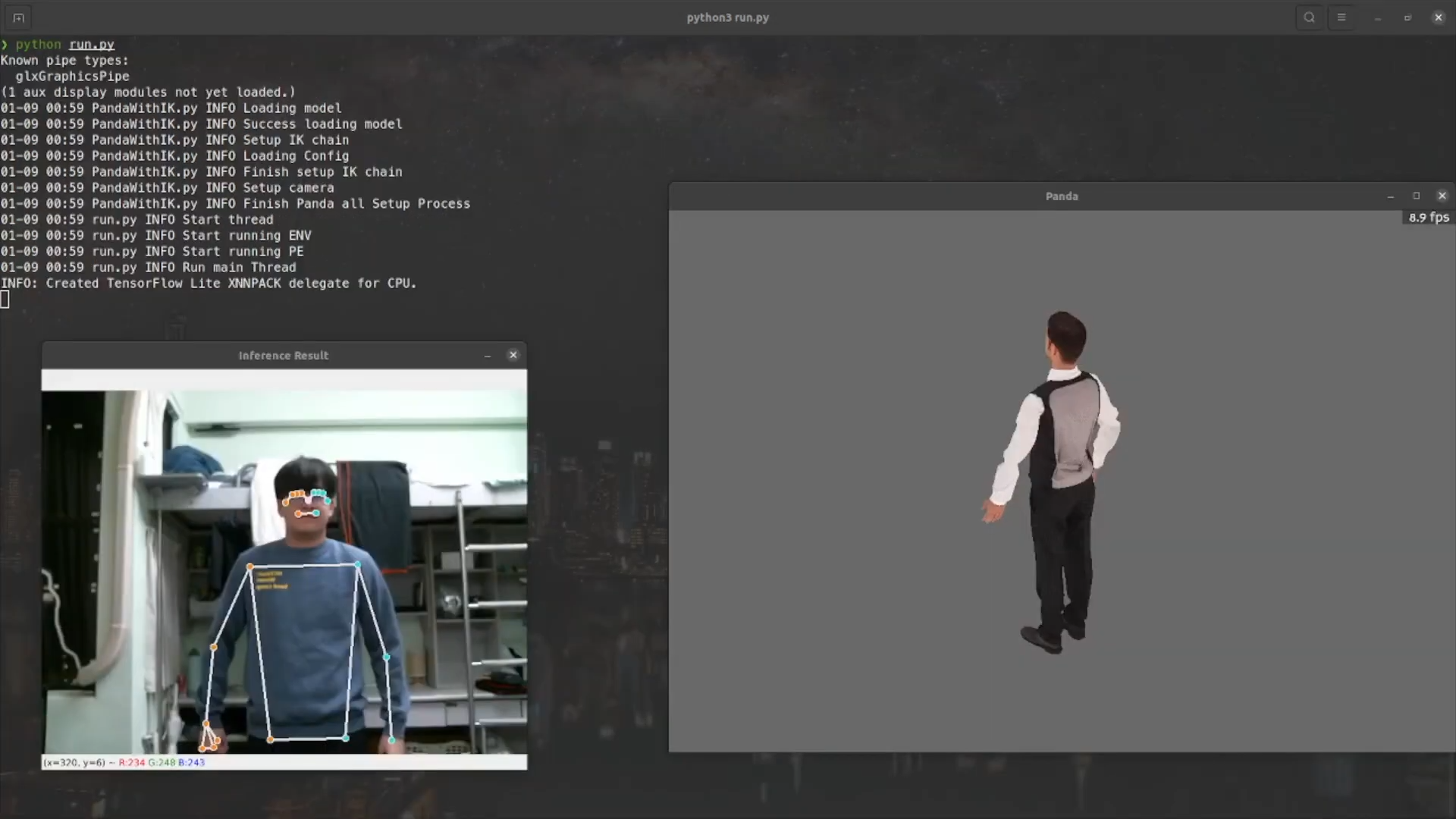}
\caption{A man standing at attention}
\label{demo_2}
\end{figure}

\begin{figure}[h]
\centering
\includegraphics[width=0.8\linewidth]{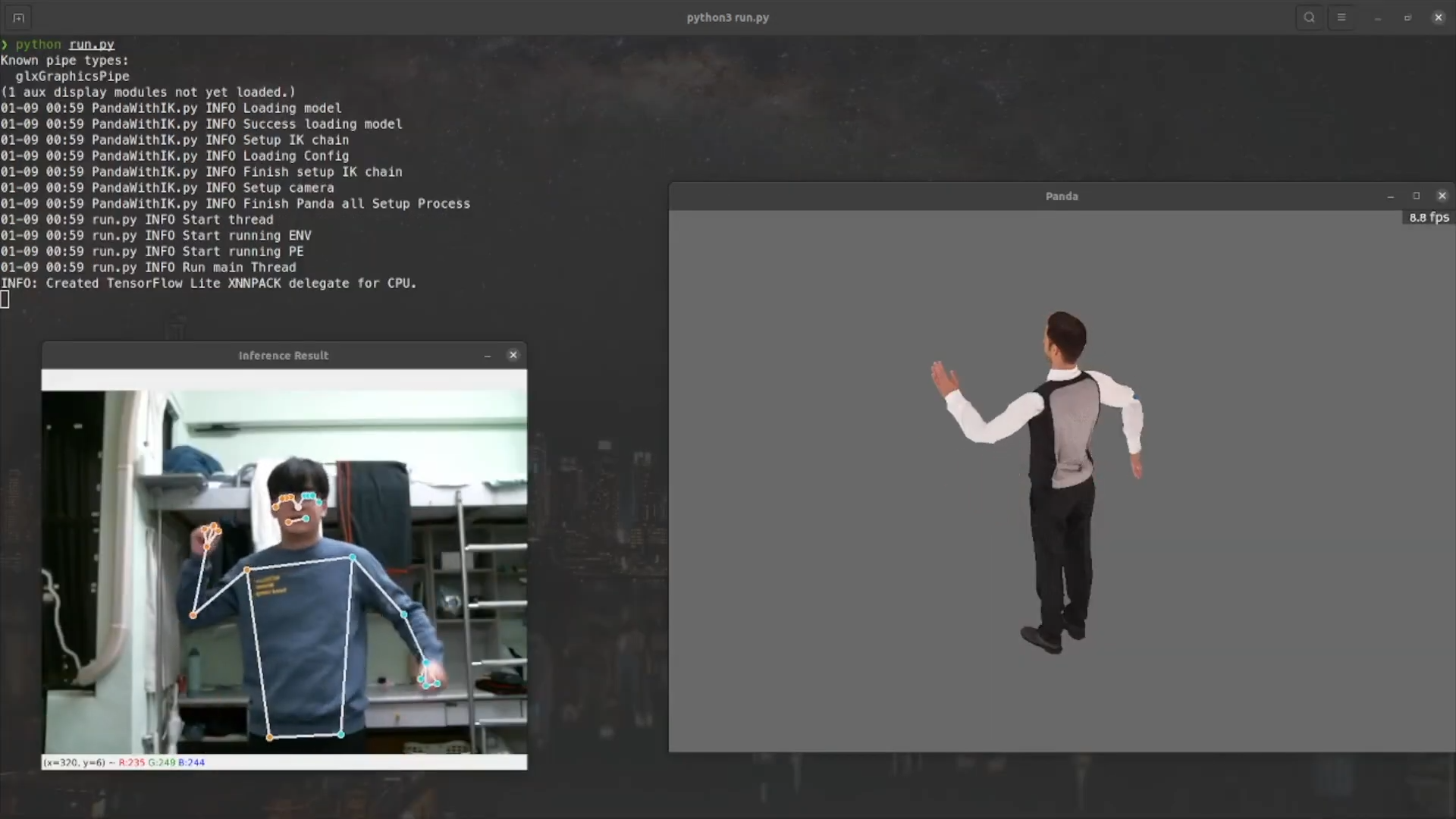}
\caption{A man raising his hand}
\label{demo_6}
\end{figure}

\begin{figure}
\centering
\includegraphics[width=0.8\linewidth]{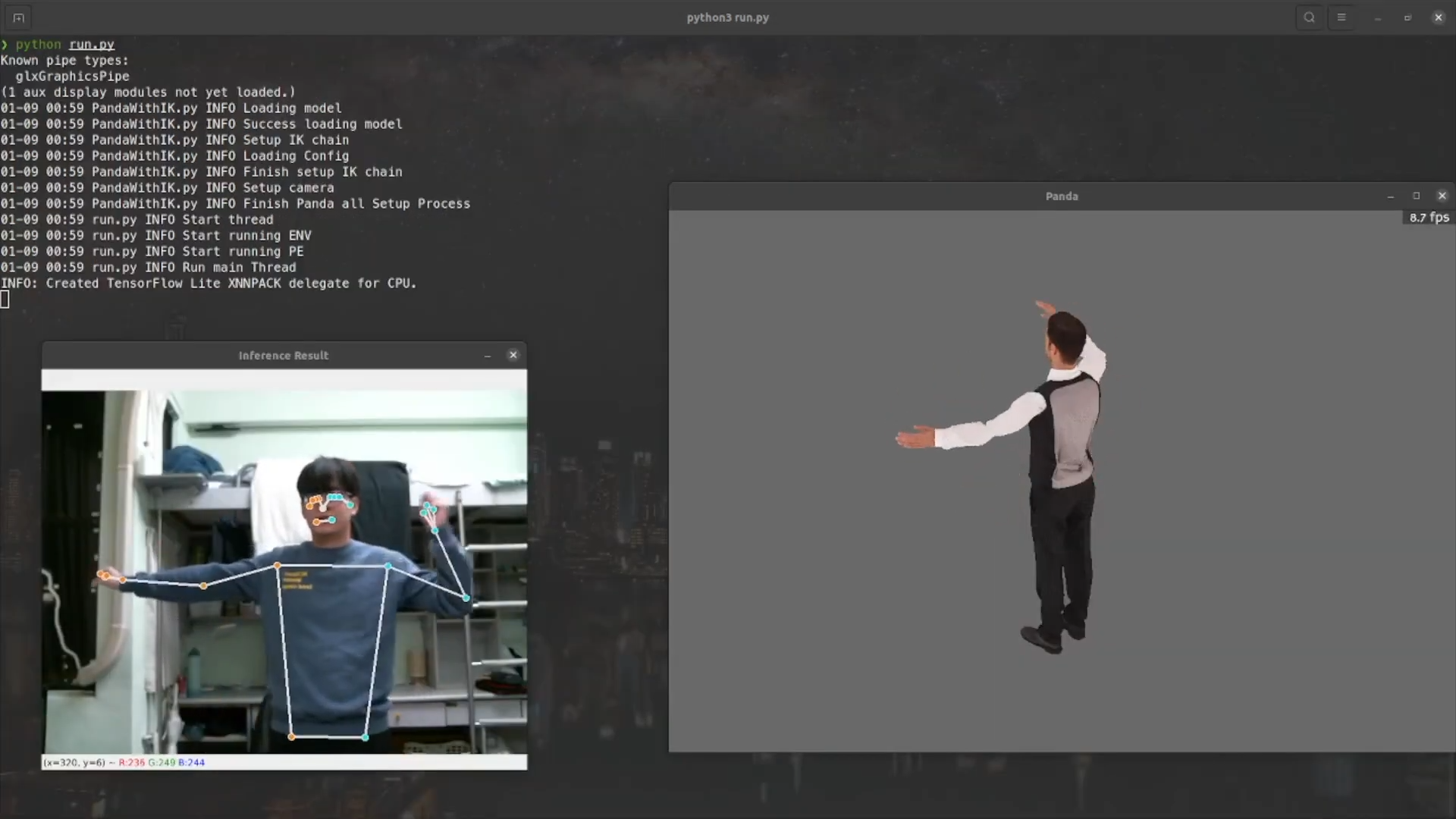}
\caption{A man directing traffic}
\label{demo_3}
\end{figure}

\begin{figure}
\centering
\includegraphics[width=0.8\linewidth]{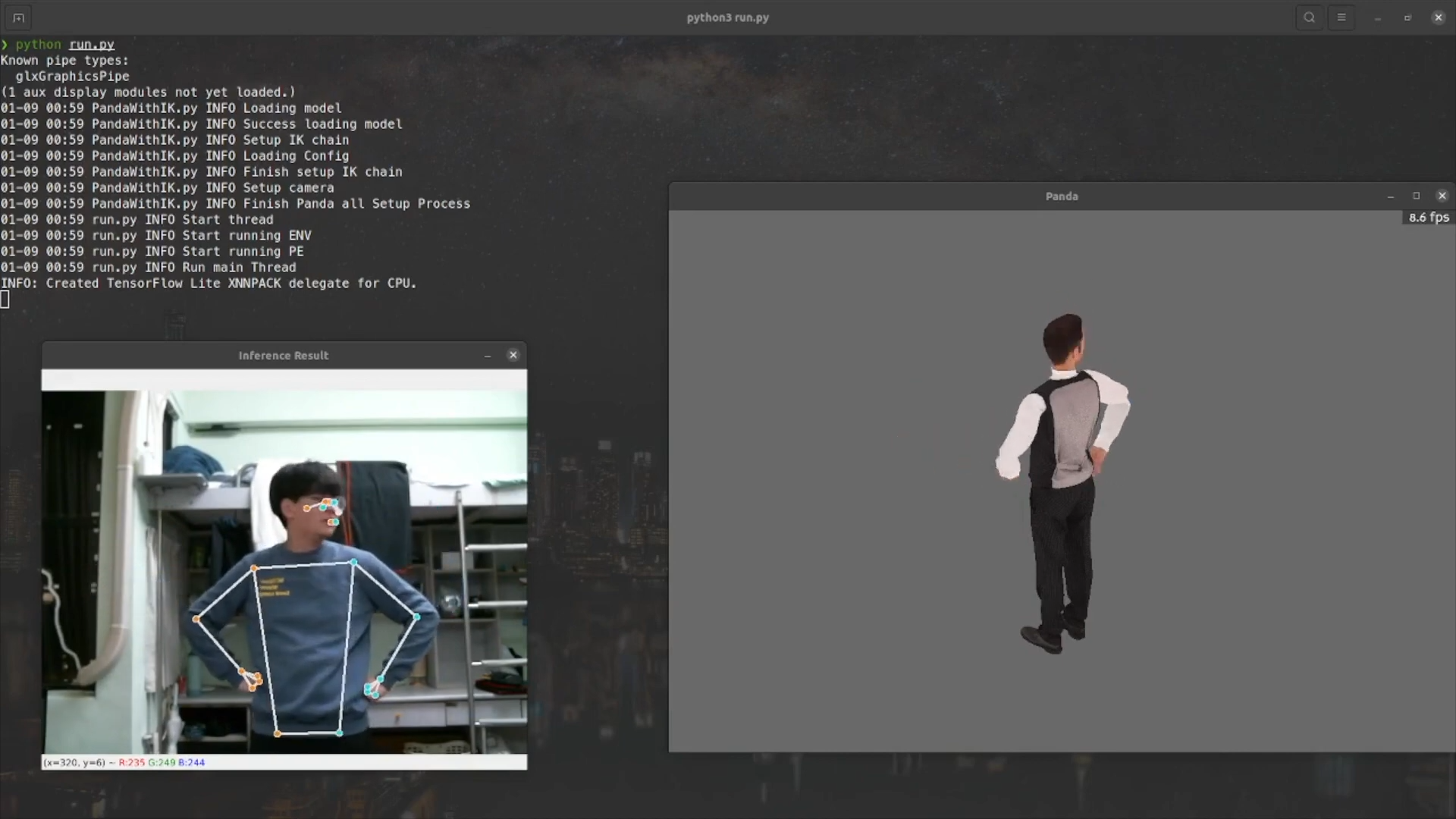}
\caption{A man hands on waist, looking sideways}
\label{demo_4}
\end{figure}

\begin{figure}[h]
\centering
\includegraphics[width=0.8\linewidth]{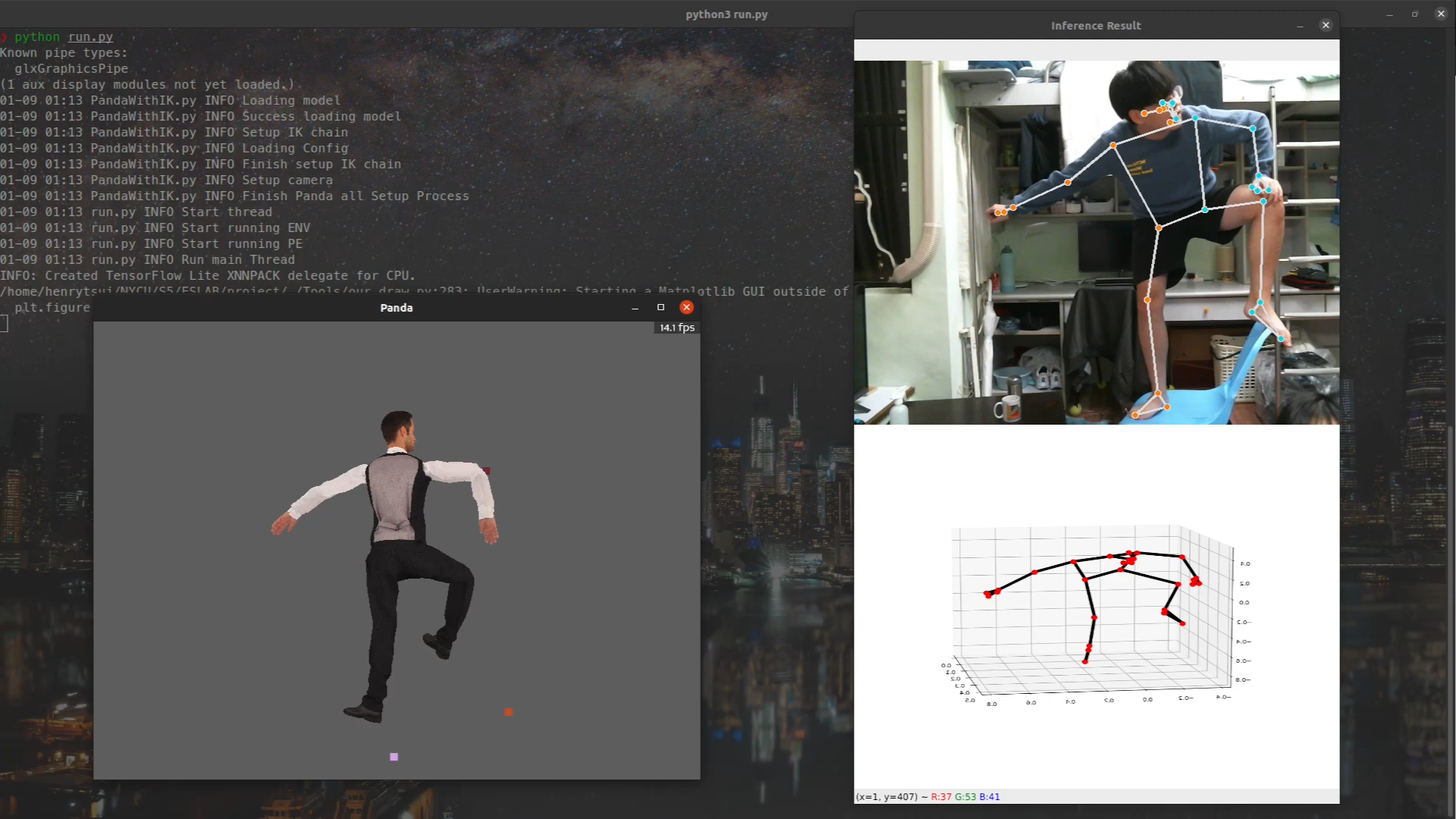}
\caption{A man hands on knee, standing on a chair, looking sideways}
\label{demo_1}
\end{figure}

\begin{figure}
\centering
\includegraphics[width=0.8\linewidth]{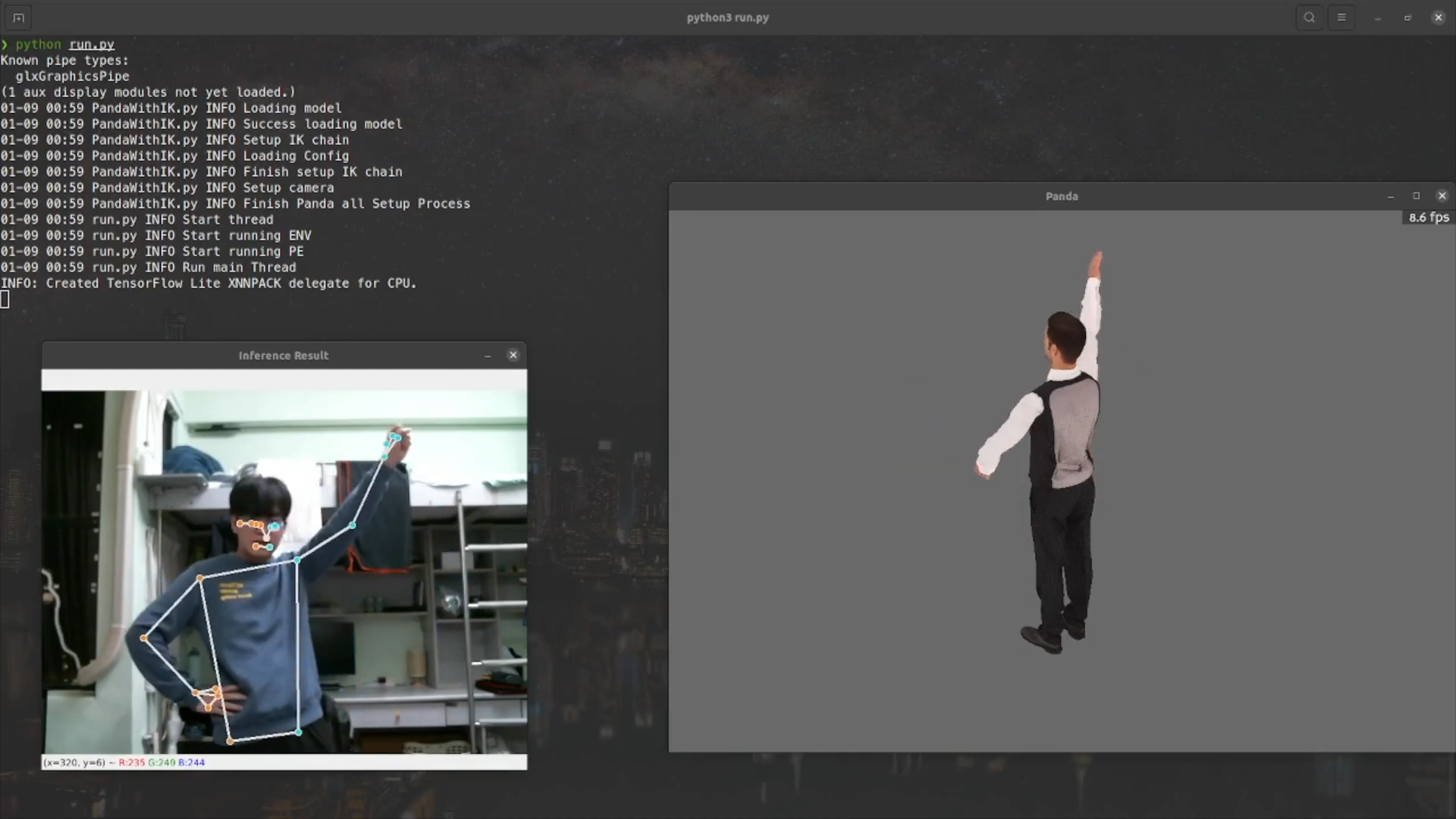}
\caption{A man doing superman pose}
\label{demo_5}
\end{figure}

\end{document}